\documentclass[a4paper,fleqn]{cas-dc}

\usepackage[authoryear,longnamesfirst]{natbib}
\usepackage{times}
\usepackage{helvet}
\usepackage{courier}
\usepackage{graphicx}
\usepackage{color}
\usepackage{amssymb}
\usepackage{amsmath}
\usepackage{multirow}
\usepackage{booktabs}
\usepackage{url}
\usepackage{ulem}
\usepackage{color,framed}

\usepackage[justification=centering]{caption} 

\def\tsc#1{\csdef{#1}{\textsc{\lowercase{#1}}\xspace}}
\tsc{WGM}
\tsc{QE}
\tsc{EP}
\tsc{PMS}
\tsc{BEC}
\tsc{DE}

\begin{document}
\begin{sloppypar}
\let\WriteBookmarks\relax
\def\floatpagepagefraction{1}
\def\textpagefraction{.001}

\shorttitle{Reasoning Chain Based Adversarial Attack for Multi-hop Question Answering}

\shortauthors{Jiayu Ding et~al.}

\title [mode = title]{Reasoning Chain Based Adversarial Attack for Multi-hop Question Answering}                      



%
\author{Jiayu Ding}



\ead{jyding20@fudan.edu.cn}




\author{Siyuan Wang}

\ead{wangsy18@fudan.edu.cn}

\author{Qin Chen}
\ead{qchen@cs.ecnu.edu.cn}


\author{Zhongyu Wei}
\cormark[1]
\ead{zywei@fudan.edu.cn}

\cortext[cor1]{Corresponding author}

\begin{abstract}
Recent years have witnessed impressive advances in challenging multi-hop QA tasks. However, these QA models may fail when faced with some disturbance in the input text and their interpretability for conducting multi-hop reasoning remains uncertain. Previous adversarial attack works usually edit the whole question sentence, which has limited effect on testing the entity-based multi-hop inference ability. In this paper, we propose a multi-hop reasoning chain based adversarial attack method. We formulate the multi-hop reasoning chains starting from the query entity to the answer entity in the constructed graph, which allows us to align the question to each reasoning hop and thus attack any hop. We categorize the questions into different reasoning types and adversarially modify part of the question corresponding to the selected reasoning hop to generate the distracting sentence. We test our adversarial scheme on three QA models on HotpotQA dataset. The results demonstrate significant performance reduction on both answer and supporting facts prediction, verifying the effectiveness of our reasoning chain based attack method for multi-hop reasoning models and the vulnerability of them. Our adversarial re-training further improves the performance and robustness of these models.
\end{abstract}




\maketitle

\section{Introduction}

Multi-hop Question Answering (QA) is a challenging task arousing extensive attention in NLP field. Compared to traditional single-hop QA tasks~\citep{rajpurkar-etal-2016-squad,rajpurkar-etal-2018-know} where the answer can be derived by simply matching the question with one single context, multi-hop QA ~\citep{welbl-etal-2018-constructing,talmor-berant-2018-web,yang-etal-2018-hotpotqa,2019QASC} requires aggregating multiple facts from different context and make composite inference across them to find the answer. 
Many recent approaches~\citep{qiu-etal-2019-dynamically,de-cao-etal-2019-question,fang-etal-2020-hierarchical} have been proposed in addressing multi-hop QA by inducing reasoning chains and claim that they perform interpretable multi-step reasoning, continuously improving the state-of-the-art performance.

\begin{figure}[ht]
    \includegraphics[width=0.475\textwidth]{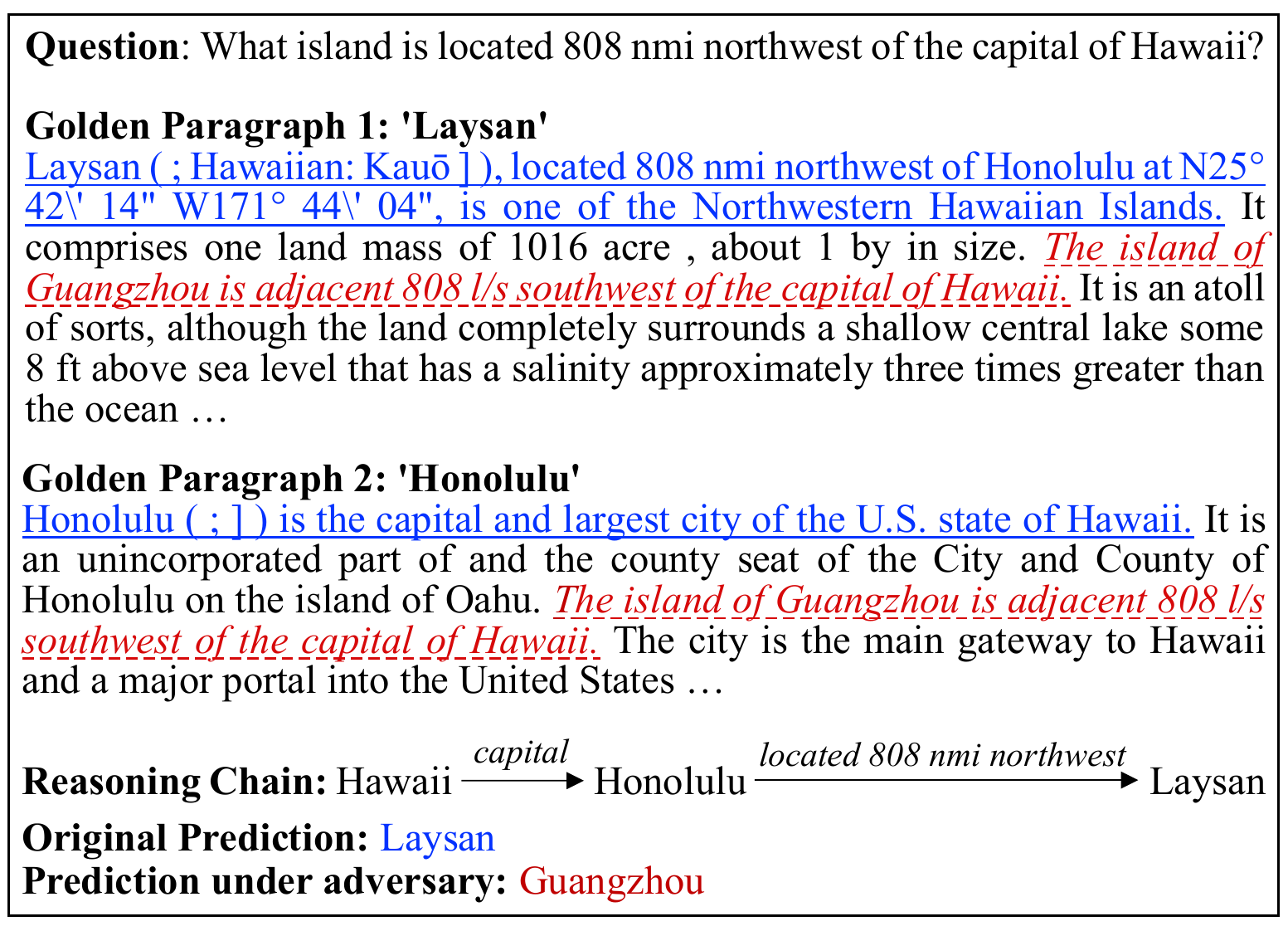}
    \caption {An example from HotpotQA dataset. The sentences shown in \textcolor[RGB]{15,68,240}{\uline{blue}} are supporting facts needed to get a correct answer. The one shown in \textcolor[RGB]{209,48,53}{\dashuline{\textit{red}}} is an artificial distracting sentence. When the distractor is inserted into original paragraphs, the QA model changes its answer from the correct \textcolor[RGB]{15,68,240}{Laysan} to a wrong \textcolor[RGB]{209,48,53}{Guangzhou}.}
    \label{a HotpotQA example}
\end{figure}

\begin{figure*}[ht]
    \centering
    \includegraphics[width=1\textwidth]{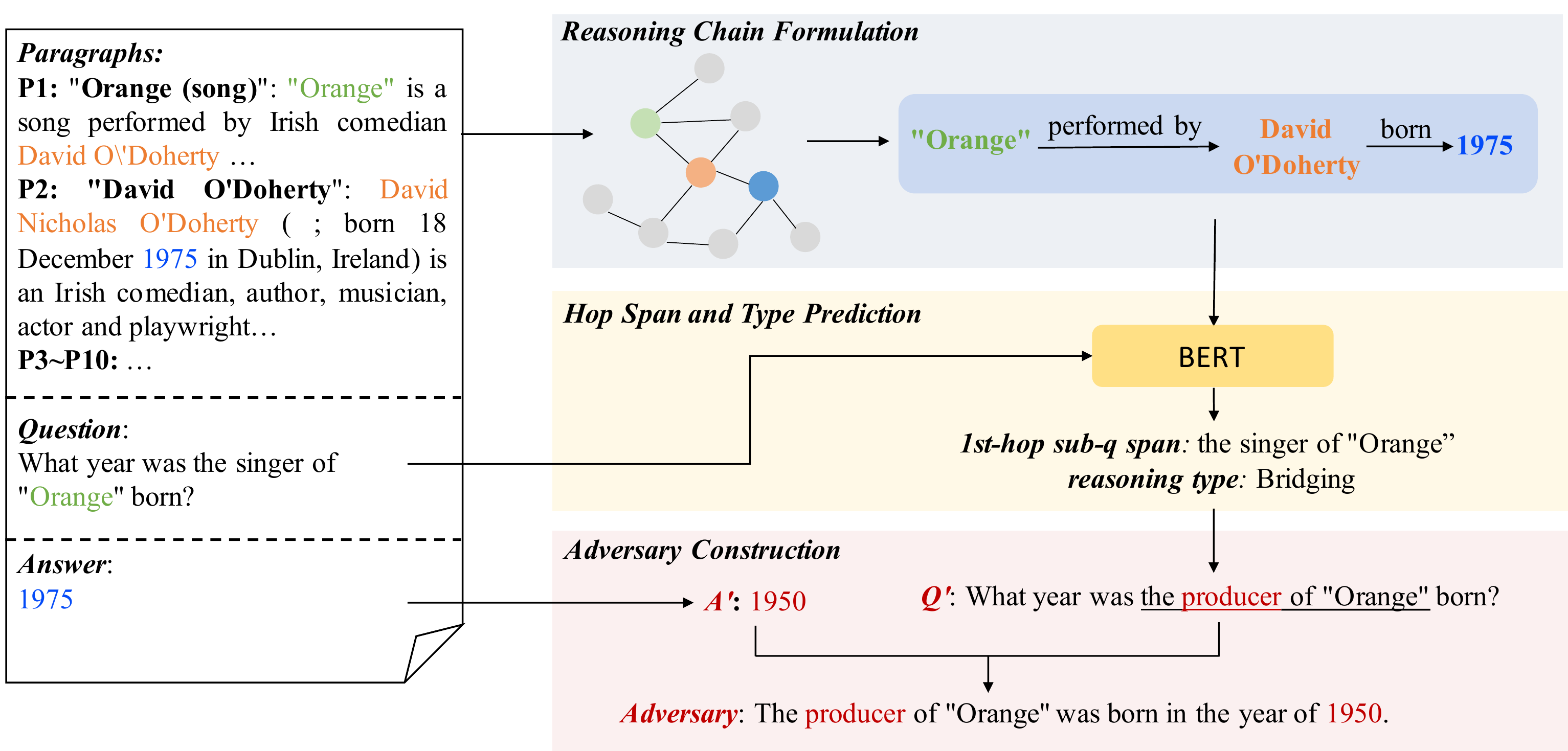}
    \caption {\small{The overall adversarial attack framework.}}
    \label{overall framework}
\end{figure*}

However, existing evaluation simply measures the standard metrics of the answer prediction, but is not sufficient or oriented for testing whether a model truly performs multi-hop reasoning. Models can achieve high accuracy by directly locating the sentences that have some keywords overlapped with the question to derive the answer, or exploiting specific biases and artifacts (e.g. answer type). These shortcut solutions allow to skip necessary reasoning hops and directly reach the answer. An example from HotpotQA dataset~\citep{yang-etal-2018-hotpotqa} is shown in Fig \ref{a HotpotQA example}. Relevant facts marked in blue are indispensable to answer the question. But one may find shortcut to the answer "\textit{Laysan}" by narrowing the prediction to "\textit{island}" and selecting the one co-occurring with  keywords "\textit{808 nmi northwest}" and "\textit{Hawaii}", in which case the second evidence sentence is actually not utilized.
Such supposition can be verified by following practice shown in Fig \ref{a HotpotQA example}.
We intentionally design a question-irrelevant but question-similar sentence (marked in red) and integrate it into the paragraphs, then the model easily breaks down and predicts a wrong answer "\textit{Guangzhou}".
The failure implies the \textit{over-stability} of the model, which means falling into text pitfalls due to relying on spurious lexical patterns without comprehensive understanding and reasoning.

Previous attempts construct adversary mainly by making modifications on the whole question via substituting entities and appending the new sentence to the input context~\citep{jia-liang-2017-adversarial,wang-bansal-2018-robust}. But these methods are not targeted for multi-hop QA models where entities play a crucial role in connecting different context to investigate reasoning chain.
Replacing entities may make the adversary quite irrelevant to the question,
thus posing limited distracting influence. 
In addition, this practice is unable to make the answer prediction traceable and identify in which link the model breaks down. 
In this work, we propose a reasoning chain based adversarial attack to inspect multi-hop reasoning ability.
Specifically, the reasoning process starting from the query entity and performing step-by-step inference by mapping queried property can be formulated as a reasoning chain, such as $\text{\textit{Hawaii}}\xrightarrow{\text{\textit{capital}}}\text{\textit{Honolulu}}\xrightarrow{\text{\textit{located 808 nmi northwest}}} \text {\textit{Laysan}}$ illustrated in Fig \ref{a HotpotQA example}.
We alter the relations instead of entities to ensure that our added adversary will not be isolated from the original context, and only change part of the reasoning chain corresponding to one hop to make the generated sentence intuitively more distracting.
As there are multiple hops, our method also allows to attack on different hops to dive into which hop may lead to a wrong prediction. 
In the above example, we attack the second hop by adding an edge "\textit{adjacent 808 l/s southwest}" pointing to a new node "\textit{Guangzhou}" (a fake answer), resulting in the adversary shown in Fig~\ref{a HotpotQA example}.

We build our reasoning chain based adversarial evaluation on HotpotQA dataset. As HotpotQA integrates different reasoning types and the corresponding reasoning chains tend to exhibit different characteristics, we customize particular attack strategies for different types.
The overall adversarial evaluation framework is designed as three steps.
First, we construct the entity-relation graph over relevant paragraphs to formulate the involved reasoning chain. Considering that the adversary which looks more similar to the question appears more effective in confusing models, we map the whole question into sub-spans corresponding to different hops of the reasoning chain for subsequent hop span based attacking. 
Finally an adversarial sentence is constructed by attacking different hop spans of the question according to different reasoning types, 
carefully altering relational phrases while keeping semantic and syntactical rationality to ensure the naturality.

We conduct our adversarial evaluation on the HotpotQA Baseline model~\citep{yang-etal-2018-hotpotqa}, DFGN~\citep{qiu-etal-2019-dynamically} and SAE~\citep{DBLP:conf/aaai/TuHW0HZ20}. The performance on answer prediction and supporting facts prediction both decrease significantly, suggesting the vulnerability of the models and underlying superficial shortcuts. 
After retraining on different adversarially constructed datasets, the models enhance robustness against such attack, and their performance on regular dataset can also be slightly improved or remain comparable. In general, we hope our insights can help promote more robust models that possess real multi-hop reasoning ability.

\section{Methods}

\begin{figure*}
    \centering
    \includegraphics[width=1\textwidth]{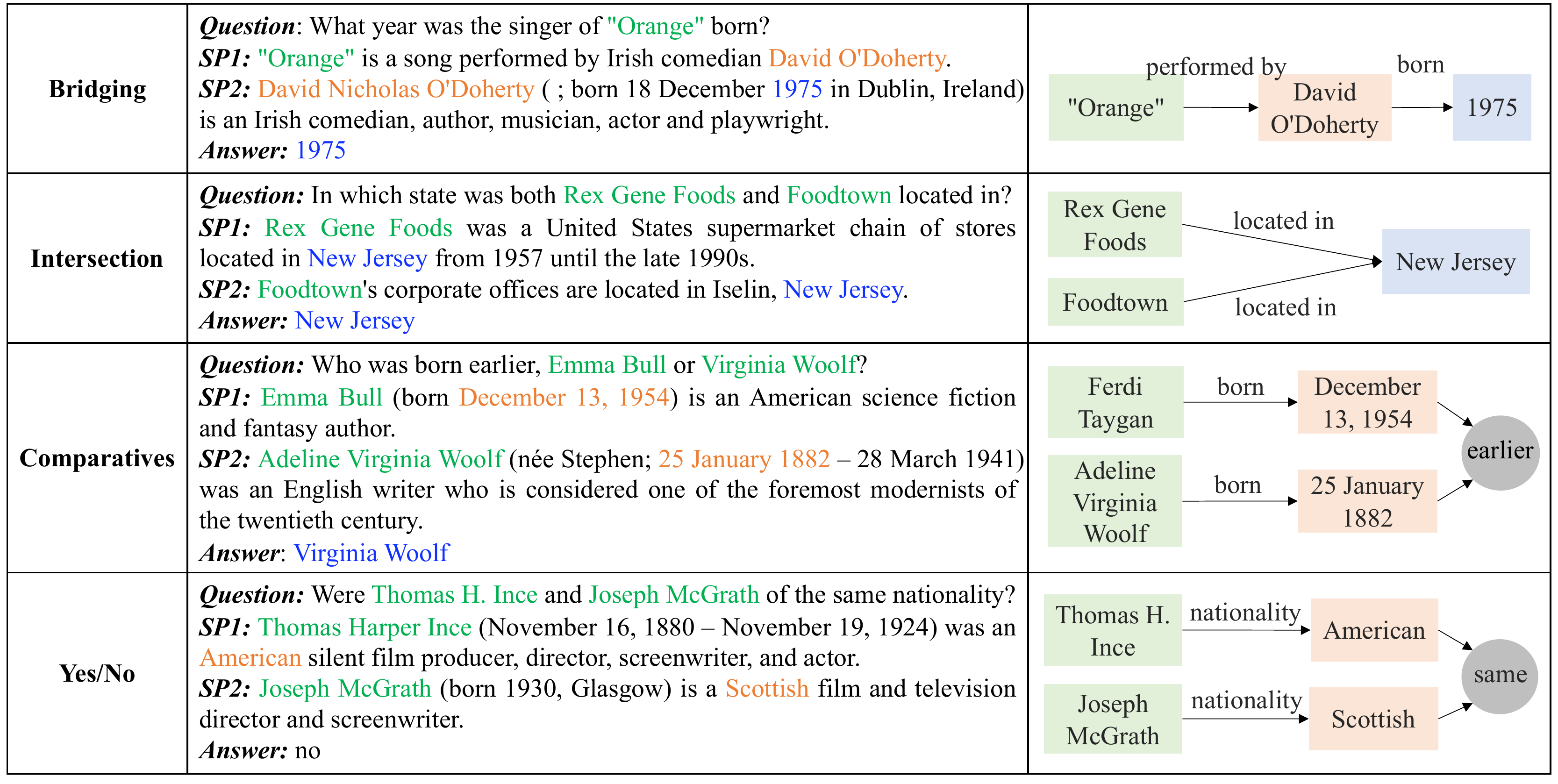}
    \caption{Examples of different reasoning types and corresponding reasoning chains in HotpotQA. The words shown in \textcolor[RGB]{78,172,91}{green},
    \textcolor[RGB]{223,130,68}{orange} and \textcolor{blue}{blue} are query entities, bridge entities and answer entities, respectively.}
    \label{reasoning types}
\end{figure*}

In this section, we introduce our reasoning chain based method to generate an adversary and the overall framework is shown in Fig~\ref{overall framework}.
In section 2.1, we construct a graph on selected paragraphs and search out the reasoning chain from query entity to answer entity along it. We also define different reasoning types according to the patterns of reasoning chain.
In section 2.2, in order to design targeted attack, we utilize the reasoning chain to automatically identify a question's type and partition the question to different segments aligning to each hop.
When the target hop to be perturbed is selected, an adversarial sentence is generated in special design by synthesizing a relational-altered question and a fake answer, as described in section 2.3.

\subsection{Reasoning Chain Formulation}

In cross-document QA tasks, the input context may include many irrelevant paragraphs which introduces a large number of entities and triples, making the graph too large to develop efficient reasoning chain searching. 
Therefore, we use a paragraph filtering module to discard unrelated paragraphs.
The concatenation of the question and each paragraph is fed into a pre-trained BERT\citep{devlin-etal-2019-bert} model followed by a classification layer, outputting a probability score to indicate the relevance, or more specifically, whether a supporting fact is in the paragraph. The ones with relevance score higher than a pre-defined threshold are selected for following graph construction.

For every example, we try to construct an entity graph over all selected paragraphs. First we use the Stanford Corenlp toolkit~\citep{2014The} to recognize named entities from the question ($\mathcal{E}_q$) and input context ($\mathcal{E}_c$). Meanwhile, we extract argument-relation-argument triples using OpenIE5 toolkit\footnote{\url{https://github.com/dair-iitd/OpenIE-standalone}} for every sentence. 
The extracted relation could be a noun-mediated phrase or a verb phrase, corresponding to property relation and action relation respectively. We re-extract from the arguments and relations if they are too long in order to get more underlying triples and clean entities. 
The arguments are further modified to align to the named entities in pre-extracted sets ($\mathcal{E}_q$ and $\mathcal{E}_c$) if applicable.  
If the argument is pronoun, we intuitively replace it with the subject entity of the previous sentence.

The whole reasoning chain is formed by acquiring the shortest paths from every query entity to the answer entity (if applicable) along the constructed graph. 
The chain actually encodes the reasoning mechanism which can be utilized to design specific attack strategies. 
Therefore, inspired by the classification ideas of some previous works~\citep{talmor-berant-2018-web, DBLP:journals/corr/abs-2002-09758}, we categorize the multi-hop QA into four reasoning types according to the characteristic of reasoning chains.
This categorization has considerable universality since up to now, most multi-hop questions can be covered.
In Fig \ref{reasoning types}, we illustrate one example for each type.

\begin{itemize}
    \item \textbf{Bridging} type QA requires sequential reasoning. One has to first infer the bridge entity and then use it to find the second hop answer. Projected onto the graph, the reasoning chain is unidirectional, with one or more bridge nodes connecting the query entity and answer nodes. The reasoning progress is by matching relation edges step by step. 
    
    \item \textbf{Intersection} type QA requires the answer to satisfy multiple conditions simultaneously. The reasoning chain involves at least two separate paths where different query entity nodes independently point to the answer. Any entity with one path broken is not an answer.
    
    \item \textbf{Comparatives} type QA compares the property of two entities. These questions usually do not have a connected path in a general sense as the former two types. The two query entities have parallel edges pointing to their own property nodes, and subsequent operation is performed on the two property nodes to get the final answer, which is often one of the two query entities and sometimes their common property. 

    \item \textbf{Yes/No} type QA specifically asks whether the two entities has a same property, and the answer is either "yes" or "no". Similar to the \textit{Comparatives} type, a special operation node "\textit{is same?}" is introduced to constitute a connected reasoning chain.
\end{itemize}

\subsection{Hop Spans and Reasoning Type Prediction}
Inspired by \large{A}\small{DD}\large{S}\small{ENT}\normalsize~\citep{jia-liang-2017-adversarial}, introducing an adversary that looks similar to the question appears more effective in beating QA models. Therefore, we implement our attack on reasoning chain based on the question. As we intend to conduct hop-targeted attack specific to different reasoning mechanisms, it is necessary to automatically acquire reasoning type and spot sub-questions, which will guide the subsequent adversary design.

In HotpotQA dataset, the \textit{Bridging} and \textit{Intersection} type QA are both categorized as \textit{bridge}, and \textit{Comparatives} and \textit{Yes/No} are both categorized as \textit{comparison}. Utilizing these existing labels, the classification for the latter two types is quite simple, just by checking the \textit{comparison} answers whether are "yes" or "no". For the former two types, we design a model to predict the reasoning type and question spans of different hops based on the reasoning chain.

Since there may be multiple named entities in the question, we intuitively choose the one with the longest shortest-path length to the answer node. This query entity node takes most hops to the answer, thus is most likely to be the reasoning start and to contain most information. If the chain involves more than two hops, we synthesize the second and latter hops as the second hop. Formally, we define $\texttt{CHAIN}$ as $\texttt{[HOP1]}$ $\texttt{ent}_q$ $\texttt{rel}_1$ $\texttt{[HOP2]}$ $\texttt{ent}_{b1}$ $\texttt{rel}_2$ $\texttt{ent}_{b2}$ $\texttt{rel}_3...$, where $\texttt{[HOP1]}$ and $\texttt{[HOP2]}$ are two special tokens. The model takes the concatenation of question tokens and chain tokens as input sequence $S = \texttt{[CLS]}$ $\texttt{QUERY}$ $\texttt{[SEP]}$ $\texttt{CHAIN}$ $\texttt{[SEP]}$. Encode the input using a pre-trained BERT model $$U=\operatorname{BERT}(S) \in \mathbb{R}^{n \times h},$$ where $n$ and $h$ are the max input length and size of BERT hidden states, respectively.

Then we apply a query mask $M \in R^{n \times h}$ and a trainable parameter matrix $W_1 \in R^{h \times 2}$ with softmax function to compute the logits, namely the probability of every token being the start position and end position of the first-hop sub-question:
$$ \mathbf{P} = [\mathbf{P}^{start}, \mathbf{P}^{end}] = \operatorname{softmax}(UMW_1) \in \mathbb{R}^{n \times 2},$$ 

\noindent The start and end positions are inferred by maximizing the joint probability
$$\text{ind}_{start},\text{ind}_{end}=\underset{1 \leq i \leq j \leq L_q+1}{\operatorname{argmax}} \mathbf{P}^{start}_i\mathbf{P}^{end}_j ,$$ where $L_q$ is the input length of query. 

For automatical reasoning type prediction, we adding a classification layer to $\texttt{[CLS]}$ token to determine whether the reasoning chain is of \textit{Bridging} or \textit{Intersection} type, respectively. This also helps the model to better encode the reasoning chains.
$$ \mathbf{P}_{type} = \operatorname{softmax}(U_0 W_2) \in \mathbb{R}^{1 \times c},$$
The optimization is based on the joint cross-entropy losses
$$\mathcal{L}=\mathcal{L}_{\text {start }}+\mathcal{L}_{\text {end }}+\lambda \mathcal{L}_{\text {type}}$$ 
where $\lambda$ is a hyperparameter.

\subsection{Adversary Construction}

\begin{figure*}
    \centering
    \includegraphics[width=1\textwidth]{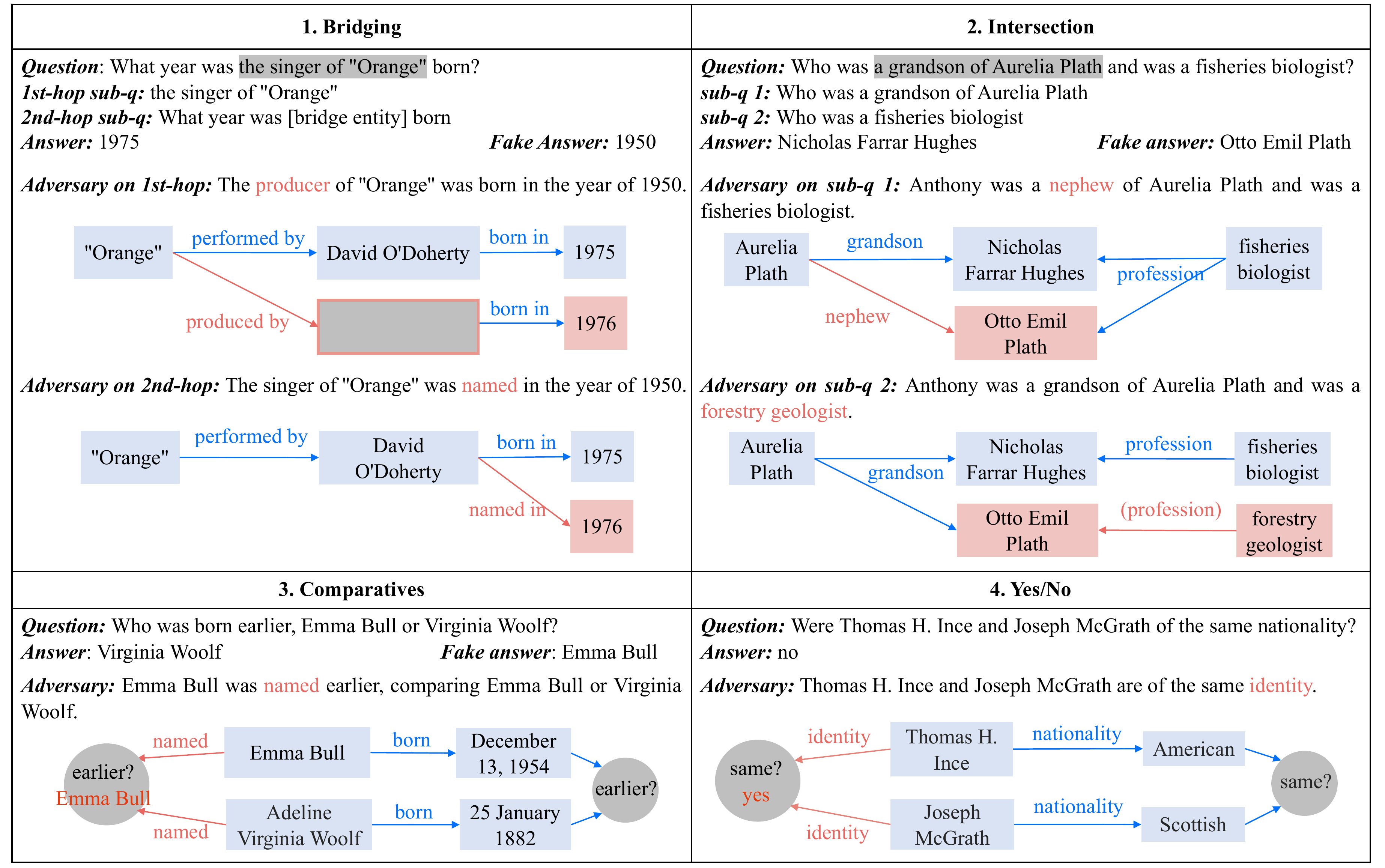}
    \caption{\small{Examples of adversarial attack according to different reasoning types and corresponding interference to the reasoning chains. The nodes and edges shown in \textcolor[RGB]{47,137,245}{blue} are gold reasoning chains, and ones shown in \textcolor[RGB]{242,116,114}{red} are the added nodes and edges introduced by the adversary. Words in shadow in the question is the hop span predicted by our model.}}
    \label{more examples}
\end{figure*}

We design specific attack tactics to different type questions. Each follows a similar procedure including modifying the question, setting a fake answer, and making incorporation.
The main inspiration is drawn from \large{A}\small{DD}\large{S}\small{ENT}\normalsize~\citep{jia-liang-2017-adversarial}.
Some examples of each type and the corresponding interference to reasoning chains are demonstrated in Fig \ref{more examples}.

\subsubsection{Bridging and Intersection} 
In \textbf{step 1}, we make a semantic-based modification on the question within the span of selected sub-question to make it ask something else. In our main experiment setting, (1) for \textit{Intersection}, we randomly choose one hop span since the two hops are equivalent in reasoning chain; and (2) for \textit{Bridging} type, we alter the first hop sub-question considering that it is more likely to bring successful attack when perturbing the reasoning at the first beginning. 
While many adversarial attack works try to add perturbations by changing entities, we propose that disturbing the relation may be a better solution. 
Generally, a relation can be noun-based property relation (e.g. \textit{the first son of}) or verb-based action relation (e.g. \textit{was born in}). Thus, our modification targets are common nouns, adjectives and verbs (except auxiliary verbs and model verbs), while named entities (including person, organization, location, number) are left unchanged. We first try to replace them with antonyms from WordNet, and if failed, change them to their nearest word in GloVe word vector space. Also, we restrict the substitution word to have the same NER type and part of speech, and do not share the same word stem with the original word. In the above examples, "\textit{the first son of}" will be altered to "\textit{the last daughter of}" and "\textit{was born in}" will be altered to "\textit{was named in}".

The new relation generated in step 1 cannot be endowed to the original answer because it may contradict to the original context. Therefore, \textbf{step 2} generates a fake answer to ensure compatibility. 
We extract named entities from all answers of training dataset and classify them into different NER types to build a fake answer set. For each non-stopword word in the answer string to be attacked, we first try to find the substitute using the method the same as the first step. If failed, randomly choose one that has the same type from the fake answer set. For example, substitute for "\textit{1998}" is "\textit{1999}" from the Glove nearby words and substitute for "\textit{Nicholas Farrar Hughes}" is "\textit{Otto Emil Plath}" from the fake answer set.

In \textbf{step 3}, we integrate the modified question and fake answer to generate the final adversary sentence. To be specific, we build the constituency parsing tree using Stanford Corenlp toolkit~\citep{2014The}, and design a set of conversion rules modified on \citep{jia-liang-2017-adversarial} work to convert a special question into a declarative form. In the running example, this gives the result "\textit{Otto Emil Plath was a nephew of Aurelia Plath and was a fisheries biologist.}" 
The generated sentence is quite similar to the original question, in which one sub-question totally overlaps and another is slightly modified with the syntactic structure unchanged. Obviously "\textit{Otto Emil Plath}" is not a valid answer as it satisfies only part of the original question.
The generated adversary sentence is inserted to random positions of each paragraph independently, and the supporting facts label is revised if needed.

\subsubsection{Comparatives and Yes/No} 

\textit{Comparisons} have a quite different reasoning mechanism which compares the property of two entities. There are no explicit multi hops in the question, thus requiring a slightly different altering strategy.
Considering there is a special comparing operation node in the reasoning chain, we propose to add a conceptual path which directly performs the same operation.

First, we change the queried property relation similar to the practice for \textit{Bridging} and \textit{Intersection}.
Meanwhile, we build a constituency parsing tree and identify the two competitors by extracting two arguments preceding and following the word "\textit{and}" or "\textit{or}". To get a more likely attack, (1) for \textit{Comparatives}, we deliberately set the one other than the true answer as the fake answer; and (2) for \textit{Yes/No} type, we construct affirmative sentences for those with original answer "yes" and negative sentences otherwise. 
The last step is also to convert the altered question into a syntactically correct statement.

\section{Experiments}
\subsection{Experimental Setup}
We evaluate our adversarial attack method on the dev set of \textbf{HotpotQA} dataset~\citep{yang-etal-2018-hotpotqa} in \textit{distractor} setting which has 7,405 examples. HotpotQA is currently the most widely used benchmark for multi-hop QA. It is text-based with questions expressed in natural language, and it covers all the four reasoning types, making it most suitable and challenging for our evaluation.

The original task sets several metrics including EM and F1 scores of answer, supporting facts and joint prediction. 
We attack three models: (1) \textbf{Baseline} of HotpotQA~\citep{yang-etal-2018-hotpotqa}, which is a RNN-based model synthesizing character-level models, self-attention and bi-attention; (2) \textbf{DFGN}~\citep{qiu-etal-2019-dynamically}, a representative and fundamental work claiming to conduct explainable multi-step inference, which uses a two-hop-based fusion block to dynamically select sub-graphs and claims to perform human-like step-by-step reasoning along the entity graph; (3) \textbf{SAE}~\citep{DBLP:conf/aaai/TuHW0HZ20},  the strongest model accessible on the HotpotQA leaderboard~\footnote{\url{https://hotpotqa.github.io/}}, which is GNN-based with contextual sentence embeddings as nodes instead of using entities as nodes.

\subsection{Implementation Details}
The paragraph filtering module is trained on HotpotQA training set with "supporting facts” ground truth as supervision label. The model achieves 97.1\% recall for all gold paragraphs and reduces the number of paragraphs to less than four for 85.5\% of the dev set examples, ensuring the feasibility of answer searching as well as reducing the scale of subsequent graph construction to great extent. We use BERT-base uncased pre-trained model for our sub-question and reasoning type prediction of \textit{Bridging} and \textit{Intersection} QAs. The maximum input sequence length is set to 150 and too long reasoning chains are truncated. We use Adam optimizer with learning rate of 5e-5. For training, we randomly select 700 and 200 examples which are labelled as \textit{bridge} type in the original dataset from training set and dev set respectively. The others with \textit{comparison} label are regarded as either \textit{Comparatives} or \textit{Yes/No} QAs. Then we manually annotate the sub-question's start position, end position and reasoning type for each question. After trained on these 900 examples, our model achieves 72\% F1 on joint prediction of sub-question span and 92\% accuracy on reasoning type.

\subsection{Main Results}

\begin{table*}[th!]
\caption{Main results of Baseline model, DFGN and SAE on the original HotpotQA dev set and our adversarially constructed dev set.}
\begin{center}
\begin{tabular}{cccccccc}
\toprule
\multicolumn{2}{c}{} & Ans EM & Ans F1 & Supp EM & Supp F1 & Joint EM & Joint F1 \\ \midrule
\multirow{2}{*}{Baseline} & ori & 42.5 & 56.7 & 16.5 & 59.2 & 8.3 & 35.6\\
 & adv & 29.9 (-12.6) & 41.4 (-15.3) & 1.4 (-15.1) & 19.9 (-39.3) & 0.7 (-7.6) & 9.8 (-25.8)\\ \midrule
\multirow{2}{*}{DFGN} & ori & 55.9  & 69.7 & 52.9 & 82.1 & 33.9 & 60.2 \\
& adv & \multicolumn{1}{l}{\textbf{25.6 (-30.3)}} & \multicolumn{1}{l}{34.6 (-35.1)} & \multicolumn{1}{l}{4.9 (-48.0)} & \multicolumn{1}{l}{27.4 (-54.7)} & \multicolumn{1}{l}{3.0 (-30.9)} & \multicolumn{1}{l}{14.1 (-46.1)} \\ \midrule
\multirow{2}{*}{SAE} & ori & 67.7  & 80.7 & 63.3 & 87.4 & 46.8 & 72.7 \\
& adv & \multicolumn{1}{l}{43.0 (-24.7)} & \multicolumn{1}{l}{54.1 (-26.6)} & \multicolumn{1}{l}{26.6 (-36.7)} & \multicolumn{1}{l}{54.1 (-33.3)} & \multicolumn{1}{l}{19.1 (-27.7)} & \multicolumn{1}{l}{39.3 (-33.4)} \\ 
\bottomrule
\end{tabular}
\label{main results}
\end{center}
\end{table*}

The overall performance of the three models on the original dev set and our adversarial dev set are shown in Table \ref{main results}. All the models show significant performance decrease in answer and supporting facts prediction. Especially for Baseline and DFGN, answer EM drops to below 30 points and supporting facts EM are even more incredibly low (only 1.4\% for Baseline and 4.9\% for DFGN). Though SAE is comparatively stronger, it also experiences 24.7 and 36.7 points decrease on answer EM and supporting facts EM, respectively. We investigate the examples which make wrong predictions, and find that 51.1\%  examples mistakenly take our fake answer as answer and 89.2\% take our adversary sentence as supporting fact. On the other hand, although these models almost fail on supporting facts prediction, they somehow succeed on answer prediction. This casts doubt on the interpretability of the models' multi-hop reasoning, since an answer should not be found in lack of sufficient evidences to implement complete reasoning. In a word, the results verify the effectiveness of our attack that the adversary can successfully distract the models' attention and mislead them by superficial resemblance to the question. There may exist reasoning shortcuts such as word-matching instead of developing multi-hop inference, which violates the purpose of multi-hop QA. 

Surprisingly, although DFGN and SAE outperform Baseline a lot on the original dev set, their answer EM drops more when faced with adversary (especially DFGN even performs 4.3\% lower than Baseline). Since Baseline model relies more on language understanding while DFGN and SAE utilize GNN to perform multi-step information aggregation and reasoning, it may be reasonable to deduce that our attack method poses question and challenge to more interpretable reasoning models. When altering relations on certain hops, we equivalently add distracting edges and nodes. The added chains overlap the original reasoning chains to different extent, but do not arrive the true answer. Whether the graph-based models have the ability of attending to correct relations and entities and avoiding being deviated to those broken chains is a crucial concern. 

\begin{table}[ht!]
\caption{Answer EM scores of DFGN on original dev set and adversarial dev set in different reasoning types.}
\begin{center}
\begin{tabular}{cccc}
\toprule
             & Proportion & Ori dev & Adv dev \\ \midrule
All          & 100\%      & 55.9    & 25.6    \\ \hline
Bridging     & 49.6\%     & 54.8    & 16.4    \\
Intersection & 32.0\%     & 52.2    & 26.5    \\
Comparatives & 12.9\%     & 58.0    & 37.9    \\
Yes/No       & 5.5\%      & 82.6    & 74.4    \\ 
\bottomrule
\end{tabular}
\label{resuls of different types}
\end{center}
\end{table}

We look into the detail scores of different reasoning types, and list the answer EM of DFGN in Table \ref{resuls of different types} for illustration. The detail scores of Baseline and SAE have similar tendency. \textit{Bridging} type is the most unrobust type, with EM dropping to 16.4\%. \textit{Intersection} ranks second. This testifies that our relation-based attack method is targeted to multi-hop question answering. Successive hops can get trapped to fake chains in the intermediate reasoning process and thus stray away from the true answer. On the other hand, DFGN is good at \textit{comparisons}, especially \textit{Yes/No} questions which already achieves 82.6\% on the original dev and continues to keep EM as high as 74.4\%. We explain the low attack success as that our method does not interfere or break the interested properties or gold reasoning chains, but only adds a one-hop "same" or "not same" conceptual relation, which thus has quite limited influence on the original question. 

\subsection{Comparison with other attack methods}
We compare our attack method with two previous methods. \large{A}\small{DD}\large{S}\small{ENT}\normalsize~\citep{jia-liang-2017-adversarial}  modifies the whole question by substituting all the nouns, adjectives, named entities and numbers, and combining a fake answer chosen from a pre-defined set. The modified sentence is appended to the end of the context. The drawback is that the adversary will be unrelated to the question and it is unable to identify which reasoning link breaks down. 
\large{A}\small{DD}\large{D}\small{OC}\normalsize~\citep{2019Avoiding} modifies the gold paragraph by substituting the bridge entity, which is simply taken from the paragraph title. This will fail when the test set or other datasets do not specify gold paragraphs and titles, or when the bridge entity is not the title. Furthermore, these two methods actually fail (unable to construct adversary at all) on \textit{comparison} questions. Therefore, we compare DFGN performance on \textit{bridge} ones (\textit{Bridging} + \textit{Intersection}) in Table \ref{comparison with other attack methods}, where our method brings greatest drop on answer prediction. Overall, our method has three advantages: (1) It can conduct more comprehensive attack over different question types and is not restricted by gold evidence accessibility. (2) It achieves higher attack success, which we attribute to higher relevance of adversaries to questions and the retention of entities. (3) It is able to perform targeted attack (e.g. different hops) and inspect the real multi-hop reasoning ability.

\begin{table}[ht!]
\caption{Comparison of DFGN performance with previous attack methods on \textit{Bridging} + \textit{Intersection} dev examples.}
\begin{center}
\begin{tabular}{ccccc}
\toprule
         & Ans EM   & Ans F1   & Supp EM    & Supp F1   \\ \midrule
original & 65.3     & 80.2     & 60.2       & 86.4      \\  \hline
AddSent  & 20.1     & 29.3     & \textbf{1.1}        & \textbf{24.4} \\ 
AddDoc   & 38.9     & 54.8     & 27.9       & 63.7 \\
Ours     & \textbf{19.2}     & \textbf{28.1}     & 1.4        & 26.3 \\
\bottomrule
\end{tabular}
\label{comparison with other attack methods}
\end{center}
\end{table}

\subsection{Ablation Studies}
\subsubsection{Hop Spans}
The hop span prediction procedure allows us to attack any segment of the reasoning chain in order to detect model's ability on different reasoning steps. In the main experiments, we construct our adversary by modifying the first-hop for \textit{Bridging} type and a random hop span for \textit{Intersection} type. In this part we try different modification strategies including (1) the second-hop sub-span and (2) both hops (i.e., the whole question) on DFGN. As shown in Table~\ref{resuls of different hops}, the model is less distracted when the alteration is based on both hops. It is quite understandable that the resulting adversary sentence does not overlap with the desired reasoning chain at all, thus is less confusing. 
For \textit{Bridging} type, altering the second hop has lower attack success rate than altering the first hop, implying that the model may tend to use word-matching tactic during second hop reasoning, in that when we leave the second hop span same to the question, it is more misled by our distractor.

\begin{table}[ht!]
\caption{DFGN performance on adversarial dev set when attacking on different hop spans.}
\begin{center}
\begin{tabular}{cccc}
\toprule
                              & Attack span  & Ans EM & Ans F1 \\ \midrule
\multirow{3}{*}{Bridging}     & 1st hop      & \textbf{16.4}   & \textbf{24.9}  \\
                              & 2nd hop      & 17.1   & 25.6   \\ 
                              & both hops    & 28.5   & 39.0   \\ \midrule
\multirow{2}{*}{Intersection} & random hop   & \textbf{26.5}   & \textbf{37.4}   \\
                              & both hops    & 32.2   & 44.6   \\ 
\bottomrule
\end{tabular}
\label{resuls of different hops}
\end{center}
\end{table}

\begin{table}[ht!]
\caption{Performance on adversarial dev sets with different modified words.}
\begin{center}
\begin{tabular}{ccccc}
\toprule
         & \multicolumn{2}{c}{Baseline} & \multicolumn{2}{c}{DFGN} \\\cline{2-3} \cline{4-5}
         & Ans EM  & Ans F1   & Ans EM   & Ans F1  \\ \midrule
Relation & 29.9    & 41.4     & 25.6     & 34.6    \\ 
Entity   & 30.9    & 42.2     & 26.2     & 35.3    \\
\bottomrule
\end{tabular}
\label{resuls of entity}
\end{center}
\end{table}

\subsubsection{Entities}
To testify our supposition that entity plays an important role in multi-hop QA, we conduct experiments on changing all named entities in the question instead of relational words. The results shown in Table \ref{resuls of entity} verify the supposition. Answer scores of changing entities are all higher than changing relations, indicating that discarding interested entities may make the adversary irrelevant and thus less distracting.

\begin{table}[ht!]
\caption{Answer performance on adversarial dev sets with different adversary placement positions.}
\begin{center}
\begin{tabular}{ccccccc}
\toprule
         & \multicolumn{2}{c}{Baseline} & \multicolumn{2}{c}{DFGN} & \multicolumn{2}{c}{SAE} \\
         \cline{2-3} \cline{4-5}\cline{6-7}
         &  EM  &  F1   &  EM   &  F1  &  EM   &  F1  \\ \midrule
Random   & 29.9 & 41.4  & 25.6  & 35.6 & \textbf{43.0}  & \textbf{54.1}  \\ 
First  & \textbf{29.5}    & \textbf{40.8}    & \textbf{18.8}     & \textbf{26.7}  & 45.3   & 56.6   \\
Last & 31.4    & 42.8    & 27.9     & 37.7  & 45.0 & 55.9    \\
\bottomrule
\end{tabular}
\label{resuls_of_different_positions}
\end{center}
\end{table}

\subsubsection{Adversary Placement}
We conduct experiments to study whether different insertion positions of the distractor sentence will bring different attack result.  
In contrast to \textbf{random} insertion, we also place the same adversary at the beginning (\textbf{first}) or the end (\textbf{last}) of every paragraph.
As shown in Table~\ref{resuls_of_different_positions}, SAE performs worst on random insertion, while Baseline and DFGN have poorest performance when the distractor sentence is prepended to paragraphs. Comparatively speaking, the difference between random and appending strategy is close. Our choice for random insertion is based on its relatively satisfying results as well as to avoid potential superficial exploitation such as always ignoring the first sentence.

\subsection{Adversarial Retraining}

\begin{table}[ht!]
\caption{Answer EM of models trained on original training set and augmented training sets.}
\begin{center}
\setlength{\tabcolsep}{1mm}{
\begin{tabular}{ccccc}
\toprule
                          & \begin{tabular}[c]{@{}c@{}}Training $\rightarrow$\\ Test $\downarrow$\end{tabular} & 100\% Ori & 100\% Adv  & \begin{tabular}[c]{@{}c@{}}100\% Ori \\ +25\% Adv \end{tabular} \\ \midrule
\multirow{2}{*}{Baseline} & ori dev & 42.5 & 41.0 & 43.0 \\
                          & adv dev & 29.9 & 49.0 & 41.9 \\ \hline
\multirow{2}{*}{DFGN}     & ori dev & 55.9 & 49.4 & 54.0 \\
                          & adv dev & 25.6 & 44.0 & 39.8\\
\bottomrule
\end{tabular}
}
\label{retraining}
\end{center}
\end{table}

We follow the main experiment setting to adversarially convert the whole HotpotQA training set which has 90.4k examples. Then we retrain the Baseline model and DFGN using (1) all the adversarial training data (2) original training set augmented by 25\% randomly sampled adversarial training examples (116k in total). Then we evaluate these models on both original dev set and adversarial dev set. The results are shown in Table \ref{retraining}.

After augmented retraining, the Baseline model largely improves resistance against adversarial attack from 29.9\% to 41.9\%, and the latter is even comparable to the original performance (42.5\%). The evaluation on original dev set also slightly increases by 0.5\% (from 42.5\% to 43.0\%). These results suggest the effectiveness of adversarial retraining that enhances the model's accuracy and robustness.
However, we also need to pay attention to preventing models from overfitting the adversary. As shown in the table, training on entire adversarial data will undermine the model's generalizability to the regular examples (dropping from 42.5\% to 41.0\%), even though it achieves the highest score of 49.0\% on adversarial dev set.
For DFGN, although the performance on original dev set does not reach as high, its stability under adversarial attack is also enhanced(from 25.6 to 39.8). The proportion of augmented data may be further manipulated to yield better performance.

\section{Related Work}
\subsection{Multi-hop QA} 
\normalsize For single-hop QA tasks such as SQuAD~\citep{rajpurkar-etal-2016-squad,rajpurkar-etal-2018-know}, QA models can retrieve the answer by simply matching the question with the sentences of the input single paragraph. Contrastively, for multi-hop QA, any single sentence alone is insufficient to answer the question. The models are required to combine the information from at least two documents and make reasoning over them. Several datasets have been proposed for multi-hop reading comprehension, including \large{W}\small{IKI}\large{H}\small{OP}\normalsize~\citep{welbl-etal-2018-constructing} \large{C}\small{OMPLEX}\large{W}\small{EB}\large{Q}\small{UESTION}\normalsize~\citep{talmor-berant-2018-web},  HotpotQA~\citep{yang-etal-2018-hotpotqa}, QASC~\citep{2019QASC}, etc.

Different kinds of approaches have been proposed for multi-hop QA, and many resort to graph neural network. DFGN~\citep{qiu-etal-2019-dynamically} repeatedly performs one-hop reasoning over the entity graph constructed from the context and dynamically selects relevant subgraph to propagate information at each step. KGNN~\citep{DBLP:journals/corr/abs-1911-02170} enhances the entity graph based on co-reference by adding relational edges from knowledge graph (KG). CogQA~\citep{ding-etal-2019-cognitive} gradually expands the cognitive graph by adding the next-hop entities and possible answer spans of each existing node. Different from entity graph, SAE~\citep{DBLP:conf/aaai/TuHW0HZ20} uses contextual sentence embeddings as nodes and treats supporting facts prediction task as node classification. HGN~\citep{fang-etal-2020-hierarchical} creates a hierarchical graph to synthesize information from different levels of granularity and promote interactions among them. 
Similar work includes~\citep{2018Exploring,tu-etal-2019-multi,de-cao-etal-2019-question,2020Is}. 
Other approaches incorporate memory network~\citep{2018An,2016Who}.

 \subsection{Adversarial Attack}
 While recent works have revealed great progress in multi-hop QA and keep promoting state-of-the-art performance, the true ability of machine reading comprehension and multi-fact reasoning as well as model robustness is remained questioned. Learning from the field of computer vision, adversarial attack on NLP tasks can be conducted by adding slight perturbations to the input context, and the models under attack are supposed to therefore produce wrong outputs. The perturbations should not result in conflicts with the original context or change the golden answer. 
 
 For QA task, \large{A}\small{DD}\large{S}\small{ENT}\normalsize ~\citep{jia-liang-2017-adversarial} proposes the first attack attempt by concatenating an adversarial sentence to the input context. They generates the adversarial sentence by substituting the named entities in the given question and combining a fake answer chosen from a pre-defined set. Experiments report significant performance decrease of sixteen models on the single-hop QA task SQuAD. 
 Based on \large{A}\small{DD}\large{S}\small{ENT}, \large{A}\small{DD}\large{S}\small{ENT}\large{D}\small{IVERSE}\normalsize~\citep{wang-bansal-2018-robust} makes modification by enlarging the fake answer candidates set and changing the adversary placement position in the purpose of diversifying the adversaries. They report that models adversarially retrained with such strategy can improve robustness against attack.
 T3 ~\citep{wang-etal-2020-t3} designs a tree autoencoder to embed the text with syntactic structure and semantic meaning preserved, and then applies optimization-based perturbation on both word level and sentence level. This approach can realize position-targeted attack and answer-targeted attack.
 
 Compared to single-hop QA, multi-hop QA faces another additional reasoning failure commonly referred as reasoning shortcut.
 \large{D}\small{I}\large{R}\small{E}\normalsize~\citep{trivedi-etal-2020-multihop} probes the existence of disconnected reasoning by removing part of the supporting facts from the input context, in which scenario the correct answer should not be acquired under multi-fact reasoning. 
 To avoid simple word-matching strategy as used in single-hop QA,  \large{A}\small{DD}\large{D}\small{OC}\normalsize ~\citep{2019Avoiding} replaces the true answer and bridge entity in one supporting paragraph with irrelevant mentions, and adds this whole new document to the input context. 

\section{Conclusion}
In this work, we propose a reasoning chain based adversarial attack for multi-hop QA. 
By formulating the multi-hop reasoning process with a reasoning chain, we can identify different reasoning types and customize adversary design for each type. 
Our method allows to attack any certain hop by identifying different hop spans of the question, and making modification on relational words. The hop-targeted attack can inspect models' error-prone parts during the reasoning process specifically. Three QA models under evaluation both exhibit poor performance in face of adversaries, suggesting that they are not robust enough and have limited interpretability of conducting multi-hop reasoning. Our adversarial evaluation can be utilized to improve models' performance by adversarial retraining as well as motivate new model development according to the weakness detected. 

\normalem
\bibliographystyle{cas-model2-names}
\bibliography{ref.bib}

\end{sloppypar}
\end{document}